\documentclass{ieeeconf}
\usepackage{etoolbox} % for "\patchcmd" macro
\makeatletter
\patchcmd{\ps@headings}{\rlap{\thepage}}{}{}{}
\patchcmd{\ps@headings}{\llap{\thepage}}{}{}{}
\makeatother
\AtBeginDocument{\let\autocite\cite}

\usepackage{balance}
\usepackage{url}
\usepackage{cite}
\usepackage{graphicx} % for pdf, bitmapped graphics files
\usepackage{flushend}
\usepackage{tikz}
\newcommand\copyrighttext{%
  \footnotesize \textcopyright 2024 IAS-18. Personal use of this material is permitted.
  Permission from IAS-18 must be obtained for all other uses, in any current or future
  media, including reprinting/republishing this material for advertising or promotional
  purposes, creating new collective works, for resale or redistribution to servers or
  lists, or reuse of any copyrighted component of this work in other works.}
\newcommand\copyrightnotice{%
\begin{tikzpicture}[remember picture,overlay]
\node[anchor=south,yshift=6pt] at (current page.south) {\fbox{\parbox{\dimexpr\textwidth-\fboxsep-\fboxrule\relax}{\copyrighttext}}};
\end{tikzpicture}%
}

\begin{document}
%
%\frontmatter          % for the preliminaries
%
%\pagestyle{headings}  % switches on printing of running heads
%\addtocmark{Hamiltonian Mechanics} % additional mark in the TOC

%\tableofcontents

\author{Morten Roed Frederiksen, \and Kasper Støy \\ \institute{IT-University of Copenhagen Denmark}}
\IEEEoverridecommandlockouts
\overrideIEEEmargins

\title{\LARGE \bf
  Affecta-Context: The Context-Guided Behavior Adaptation Framework
}

\author{Morten Roed Frederiksen$^{1}$, and Kasper Stoy$^{2}$
  \thanks{{$^{1}$Morten Roed Frederiksen {\tt\small mrof@itu.dk} and $^{2}$Kasper Stoy \tt\small ksty@itu.dk} are affiliated with the Computer Science Department of The IT-University of Copenhagen.}
}

\maketitle
\copyrightnotice
\begin{abstract}
This paper presents Affecta-context, a general framework to facilitate behavior adaptation for social robots. The framework uses information about the physical context to guide its behaviors in human-robot interactions. It consists of two parts: one that represents encountered contexts and one that learns to prioritize between behaviors through human-robot interactions.
As physical contexts are encountered the framework clusters them by their measured physical properties. In each context, the framework learns to prioritize between behaviors to optimize the physical attributes of the robot's behavior in line with its current environment and the preferences of the users it interacts with.
This paper illlustrates the abilities of the Affecta-context framework by enabling a robot to autonomously learn the prioritization of discrete behaviors. This was achieved by training across 72 interactions in two different physical contexts with 6 different human test participants. The paper demonstrates the trained Affecta-context framework by verifying the robot's ability to generalize over the input and to match its behaviors to a previously unvisited physical context. 
\end{abstract}

\section{Introduction}

The physical context of our interactions with fellow humans has a
significant influence on our behavior. We shape our behavior through
continuous interactions with the environment. We moderate different
aspects of our behavior with each physical context we encounter
\autocite{greengros2014}. For social robots, the physical context may
pose a challenge for them to communicate effectively
\autocite{Torre2020HowCS}. The physical intensity, the proxemics, the
audio volume, and the amplitude of movements are all examples of
adaptable parameters that can in or decrease a robot's ability to
communicate. If these are not attuned to the physical demands of the
context the robot may not be able to communicate effectively. As an
example, a loud speaking robot can be heard at all times but would
probably be a terrible fit for more intimate interactions.

When social robots are designed they are usually equipped with a set of
discrete behaviors to handle a variety of social interactions. While
such robots are well attuned to specific contexts, placing them in a
different context can make some of their actions seem inappropriate.
This could introduce a negative effect on the social capabilities of the
robot. English \& Carstensen 2014 discuss that the humans in human-robot
interactions determine the amount of affective impact from the robot's
actions. This complicates a dynamic adaption to contexts for robots
further - as every human evaluates its actions from a different
perspective, making certain actions more effect-full at certain times in
specific contexts \autocite{english2014}. Also, the social, cultural,
and physical contexts are important for the robots because we humans
interpret their actions in light of the context in which they are
performed \autocite{belpaeme2014}.

\begin{figure}[h] \centering \includegraphics[width=0.5\textwidth]{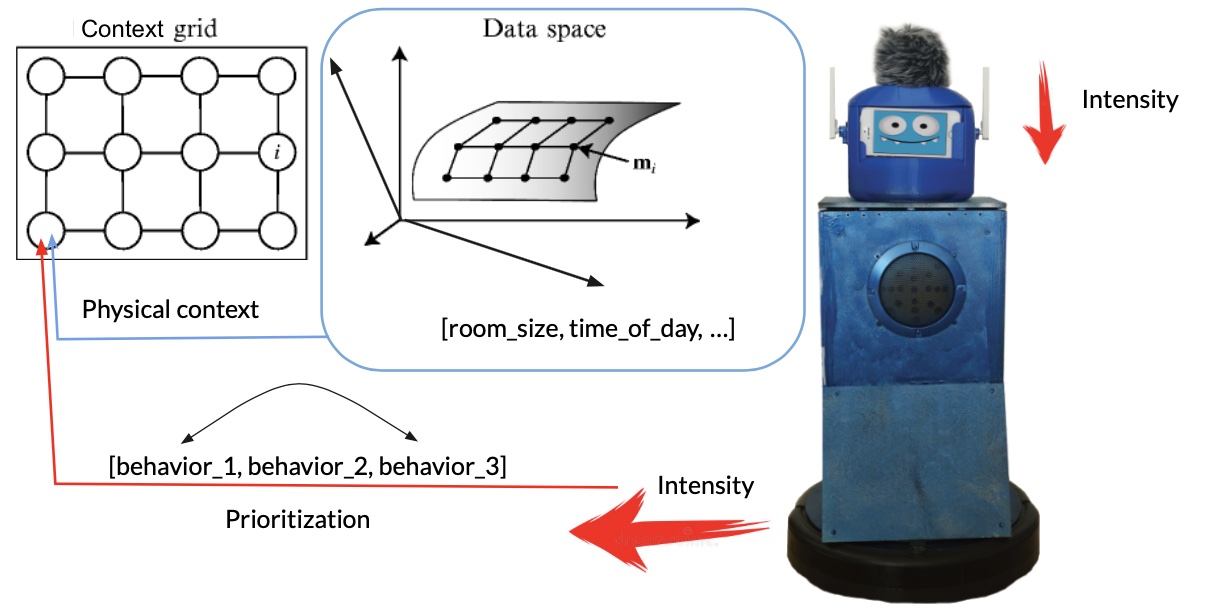} \caption{The Affecta-context framework: Each physical context is represented by areas in the context-grid (left). Each area is represented by its physical attributes such as physical proportions, terrain, etc. (middle). For each context, there is a set of behaviors ordered by how well they match the context (bottom). The Affecta-context framework guides the physical intensity of the robot ensuring it matches the robot's current context (right).}
\label{affecta}
\end{figure}

If the physical setting is known ahead of an interaction it may be
viable to pre-define the best fitting behaviors. However, It is often
not the case that robots possess sufficient information ahead of
interactions, and there is also no guarantee that the predefined
physical proportions pertaining to a specific scenario or user
preference remain valid throughout a whole interaction. This underlines
an argument for using dynamic learning strategies as Weber et al.~2018
suggest by adapting a robot's humoristic behavior to specific user
preferences \autocite{Weber2018HowTS}. Some contextual information may
also be unattainable ahead of an interaction. For instance, the expected
affective impact on humans (how much the current interaction changes the
current affective status of those that interact with the robot) may vary
with different robot morphologies, different cultural contexts, etc.
This paper argues that it may be beneficial for social robots to
continously learn the affective impact of behaviors in specific physical
contexts along with open-ended discovery of new environments and it
presents a framework to facilitate this.

\subsection{Previous approaches}

Adapting behaviors to the context has been found effective as a driver
for more diverse behavior choices in social robot projects, and
developing context-aware systems may be the hurdle to overcome to unlock
better social functionality of robots outside of lab conditions
\autocite{Luber2012,Banisetty2019SociallyAwareNA}. There have been a few
projects on optimizing robot capabilities according to the physical
context. Narayanan et al.~2011 created movements based on visual
perception of the environment and Jamone, Damas, and Santos-Victor 2014
created dynamic mapping models based on interactions with various
objects \autocite{Narayanan5979794,Jamone6967617}. The models in the
latter allowed the robot to approximate the torque rate for correctly
interacting with different context objects. Pandey et al.~2010 created a
framework that paid special attention to humans as objects in the
vicinity as their navigation system analyzed local clearance and
environment structure \autocite{Pandey5649688}. The acoustic properties
of environments were used in Lera et al 2017 to classify indoor contexts
\autocite{Lera2017ContextAI}. The project investigated using
convolutional neuronal networks to classify different contexts based on
ambient audio signals. Xiao et al.~attempted to increase the contextual
knowledge of a robot in interaction by allowing communication through
natural body language. The robot they created could understand the
meaning of human upper body gestures and would communicate by using
movements, facial expressions, and verbal language \autocite{Xiao2016}.
Robots that autonomously determine the personality traits of the users
were the focus of Zafar et al.~2018 and Zafar et al.~2019. Their
solution used speech characteristics and found promising results in
detecting personality traits
\autocite{10.1007/978-3-030-00232-9_7,8624975}. Their approach used
excerpts from the interactions annotated by a psychologist to link
nonverbal cues to a level of exhibited extroversion. A common link in
these projects is that they investigate the adaptation of robots and
robot behaviors to specific physical attributes in an environment. This
is often achieved using individual context representations tailored to
each specific task environment. This project aims to extend this
research and develop a general context-aware framework that utilizes
multiple forms of contextual information to adapt discrete behaviors to
the current physical environment. The framework we present offers
dimensionality reduction which allows it to be used with any number of
different attributes to represent each context. This allows it to be
used with both simple and extensive sensor setups.

\subsection{Context-based behavior selection}

In this paper, we present Affecta-context, a general framework that
provides context-guided behavior selection across multiple physical
contexts. The framework gradually learns to prioritize between a set of
discrete behaviors in accordance with the robot's current environment,
utilizing a combination of preconfigured behaviors and dynamic learning
strategies. Behavior selection is optimized over multiple visits to
physical contexts and is evolved as the robot explores and visits new
contexts. The framework places all visited contexts in the topography of
nodes - each representing currently known contexts. As more contexts are
visited, Affecta-context generalizes on the input and clusters similar
context nodes in the vicinity of each other. As the robot encounters
humans in each context, it interacts with them and attempts to verify
the affective impact of one of its discrete behaviors. The visible and
audible reaction of the humans that interact with it determines the
fitness of the behavior. The Affecta-context framework was added to a
robot and trained in 72 different interactions with 6 human participants
across two different physical contexts. The Affecta-context framework
enabled the robot to distinguish between the individual contexts and
created a prioritized set of behaviors as per the users' preferences.
Through the participants feedback, a high-intensity level behavior was
determined more fitting for the robot in a physically large room of
6x5m, while a lesser intensive action was found more appropriate in a
smaller room of 2x3m. A validation experiment placed the robot in a new
physical environment and had it identify the optimal matching discrete
behavior. This identified behavior was verified as being the best
fitting behavior by a group of human observers to a significant extent
(n=90, p\(<\).05) through an experiment demonstrating the optimal vs
randomly selected behaviors. Although the framework presented in this
paper simplifies complex human affective processes, the results indicate
that it can provide a good foundation for projects on non-preconfigured
affective behavior control. It is scalable in that it offers
dimensionality reduction and may be used on multiple measure-points
defining the context.

\section{System overview}

The robot we used is a 75cm tall mobile humanoid robot. It was
constructed by extending an open-source version of iRobots Roomba robot
that provided a base to build upon and movement\autocite{roomba}.
Communication with the platform was done through FTDI via a serial
connection input. The main system and context representation was running
on a raspberry PI, while a camera sensor, text to speech recognition,
and facial and body- recognition algorithms were running on an iPhone
connected to the raspberry pi through USB via the peertalk interface
\autocite{peertalk}. The phone also utilized the apple neural engine to
process the facial expression of the users to classify their current
affective state using a pre-trained convolutional network as used in
Levi 2015 \autocite{levi2015}. The phone also provided a display to show
the facial animations of the robot. This allows it to display a variety
of different predefined expressions \autocite{romo}. The robot has three
contact sensors placed at the front bumper of the robot. It also has
three low-range distance sensors, one at the front and one on either
side. For actuation, the robot uses two electric motors in a
differential drive setup allowing the robot to move forward, backward,
and to turn around its center axis. The robot can be seen in Figure
\ref{affecta}.

\subsection{Context representation}

A vital part of the Affecta-context framework is a representation of the
different contexts. In the framework, each context is represented by a
single vector that holds all attributes that define each context. As an
example, each feature vector may hold values for the estimated room
size, ceiling height, encountered obstacles, and further attributes that
can be used to distinguish the physical context. The affecta-context
framework holds a collection of multiple feature vectors, one for each
encountered context. It provides dimensionality reduction for these
context vectors and attempts to arrange them so that similar contexts
are placed near each other. The data structure we created is inspired by
a self-organizing map (SOM) which can cluster similar vectors in the
topological vicinity of each other
\autocite{Kohonen1982SelforganizedFO}. Each of these topological
positions represents a single physical context and the topological
placement of them represents small variations in the physical
attributes, meaning that a single context may take up more than one
topological position and carve out an area of the representation. The
data structure we created also allowed the framework to define
individual attribute weights (see details below) and to provide a direct
input-to-topological-position method.

Any number of physical attributes can be used to represent the context.
For simplicity in this paper, the robot in this paper would use the
physical size of the environment estimated with simple
average-time-of-drive values in each physical context. The context
representation provides dimensionality reduction like a regular
self-organizing map and with richer sensors available, multiple physical
characteristics could be sampled to give a more precise representation
of the current context in each feature vector. The context-grid at the
left side in Figure \ref{affecta} illustrates the topology of context
nodes while the dataspace depicted in the middle illustrates multiple
attributes that define a physical context. The number of attributes used
to represent the context defines how detailed the representation is.

\subsection{Updating the context representation}

The context-representation consists of 100 different physical contexts
arranged in a 10x10 matrix of individual contexts with an area of 2d
positions for each of them. The matrix is initialized with a context
vector for each position and the attribute-values of these vectors start
randomized between 0 and 1. Each time new contexts are explored and new
context-vectors are created, the most similar context-vector in the
matrix is retrieved and each measuring point in this context-vector is
updated by the difference between each value modified by a learning
rate. The distance between context-vectors is defined as the sum of the
squared difference between all context attributes of the vectors
multiplied by an attribute-specific weight modifier. As the closest
matching context-vector is updated, the nearest contexts in the
euclidean distance around it are also updated with a learning rate that
decreases (halves) with each distance step away from the center
context-vector. In a similar style as self-organizing maps, introducing
a fundamentally different new input vector automatically creates a new
region representing a context, in the topology of the 2d matrix by
altering the existing context vectors. In our implementation, the
distance between context-vectors is furthermore calculated with
attention to the importance of each of the gathered measurements in it.
Some measure points may be more important than others and could have a
greater impact on similarity when determining the distance between
contexts. To model this, each measuring point in a context vector has an
added importance modifier that multiplies the difference value.

\subsection{Behaviors in a context}

All individual contexts represented in the 2d matrix have a set of
behaviors attached. These are ordered in priority after the best fit for
the specific context. The behaviors consist of a series of movements
(forward, backward, and turning movements), head antenna gesturing
(waving pattern), facial expression animations, and audio expressions.
Each behavior differs in audio and animations but shares the same
movement and gesturing patterns at different intensity levels from 0 to
3, meaning that each of the four behaviors has a unique intensity level.
The intensity level defines the size of the physical movements and
gestures, and also determines the intensity of the animated expression.
The highest intensity level at three has the largest movement and
gesturing actions while intensity zero does not have any movements or
gesturing but instead consists solely of animations and audio
expressions. The robot prioritizes between behaviors of the context by
testing them through interactions with humans, and by gradually
adjusting its prioritization for the identified physical context of each
encounter. The behavior priority for each context is updated with the
same strategy as the physical context-vectors with the behaviors of the
neighboring contexts being updated with a decreasing learning rate as
well. The topology of the nodes may be altered when new contexts are
discovered which indirectly influence the behaviors prioritized in
similar contexts.

\section{Experimental setup}

The framework was trained in two separate phases followed by a
validation experiment to investigate the abilities of the trained
framework. During the first training phase, the robot would explore two
different phsycial contexts after which it was manually verified that
the robot was able to autonomously create discernable representations
for each of the indentified contexts. With the second training phase the
robot attempted to define the affective impact of the four discrete
behaviors. This was done through interactions with human participants.

\subsection{Training Phase One: Physical context exploration}

\begin{figure}[t] \centering \includegraphics[width=0.5\textwidth]{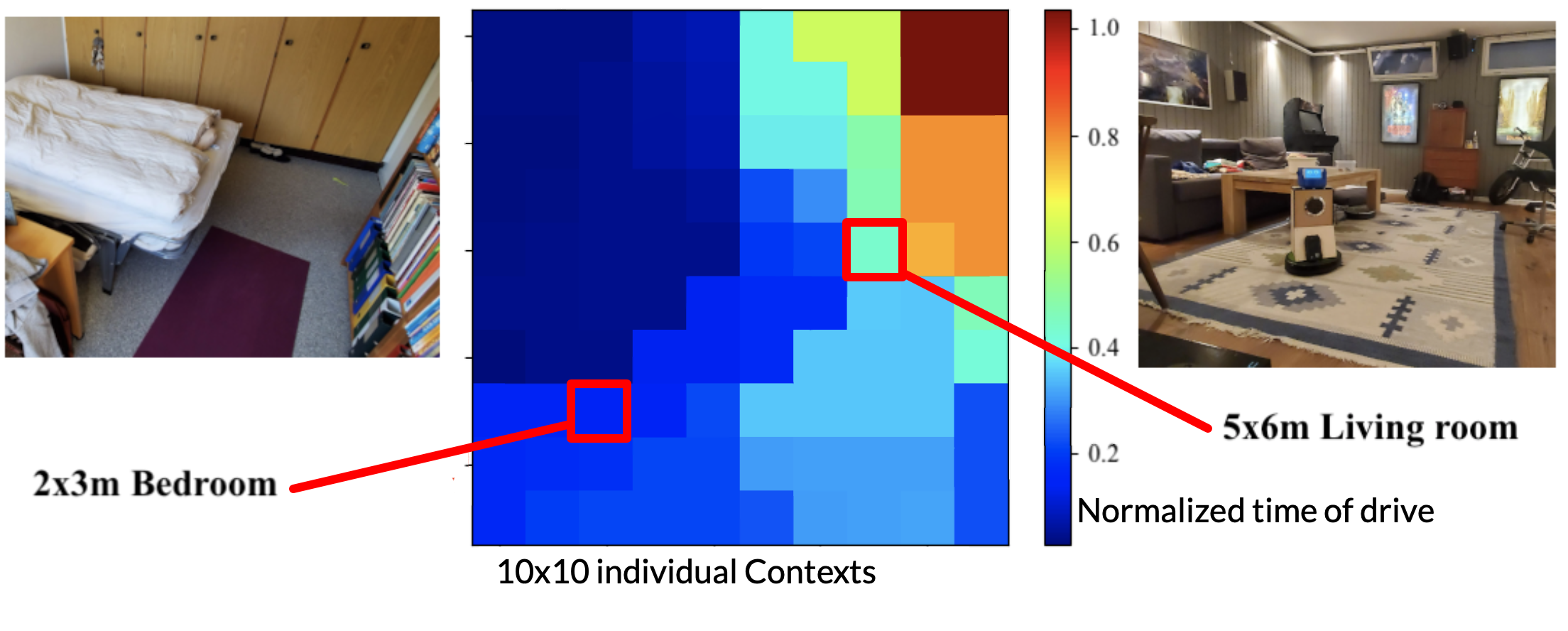} \caption{The physical context representation after 5 minutes of exploration in each context. The 10x10 squares represents the individual contexts, and each color represent the normalized time-of-drive measurement values from 0 to 1. The highlighted areas shows the individual context representation for each training context. The similar neighburhood colors shows that the framework is on the way toward creating a niche area for each of the two contexts in the representation. Note that this map shows the context representation, not a map of the physical location.}
\label{context_map}
\end{figure}

The initial training phase used the framework in two different physical
contexts, a 6x5m living room, and a 2x3m bedroom context. The robot
would randomly explore and gather measurements for a total om five
minutes in each context while simultaneously updating its context
representation. This was achieved by moving around in random directions
until the robot had performed 3 successful measurements of time-of-drive
in the physical space. The created context representation was manually
validated by expecting the visualized context-representation in relation
to the average time of drive measure point as seen in Figure
\ref{context_map}. In that visualization it can been seen that the robot
creates areas in the context-representation to represent each context
and contexts that have similar attributes (similar physical values
e.g.~physical size). The free available space may be smaller in either
of the two test locations. The framework accounts for that automatically
as it measures the available space.

\subsection{Training Phase Two: Behavior prioritization}

The second training phase focused on developing a prioritization between
behaviors in each context in regards to information gathered in
human-robot interactions. The robot would interact with 6 different test
participants (recruited with informed concent). The training sessions
were performed in the two different physical contexts. The human
participants would interact individually with the robot one after
another from a similar proximity between 1 and 2 meters. Each
interaction lasted for about ten minutes and the participants were
offered a chair to sit on if they felt like it. The participants were
informed that the robot would interact with them as soon as it could see
them. They were not instructed to perform any specific actions other
than to interact with the robot and answer truthfully to any questions
it would ask them. With each participant the robot would initiate a
conversation and ask to the participants to determine the best of two
randomly selected candidates for behaviors that would fit the current
physical context. The learning strategy for this behavior selection
across the various interactions followed a classic epsilon greedy
reinforcement approach with an exploration part and a verification part
(in a 4/1 ratio decreasing over time). The exploration path used a
random behavior while the verification strategy would use and test the
currently rated best-fitted behavior. The fitness number for each
behavior was updated following the interaction and the sum of positive
votes out of the total number of votes would define the behavior-rating
for that specific physical context going forward. During the full range
of training, the impact of 72 behaviors were investigated across the two
contexts and 96 physical context measurements were gathered.

\subsection{Validation Experiment: Autonomous behavior selection}

We performed a validation experiment to verify that the chosen
behavior-intensity of the robot in a previously unvisited context to a
significant extent was also determined to be the best fitting behavior
by human observers. The robot was given three opportunities to determine
the best possible behavior intensity for its current environment. Each
attempt included gathering three measure points and avering those to
find the best matching physical context in the trained context
representation. The changed physical context consisted of a 4 x 4m room
that was chosen for its physical proportions occupying the middle-ground
between the two previously visited physical training contexts. Although
this was an already completed physical test-setup, the current COVID-19
pandemic forced us to move this interaction to Amazons online mTurk
platform. We recruited 90 participants who each viewed a randomized
video comparison between the behavior found most fitting by the robot
and one of the other three included behaviors. With the presented videos
we tried to give the participants as good as possible view of the
physical environment and of the robot's behavior by placing the camera
in similar positions as the participants would be placed in a real
physical encounter with the robot. The participants were asked to
determine which of the two behaviors they found most fitting for the
robot's current physical environment. The participants were also also to
rate how each behavior matched the current physical context on a likert
scale from one to ten with one being ``poorly matched'' and 10 being
``perfectly matched''.

\section{Results}

\subsection{Manual verification of the training phases}

\begin{figure}[t] \centering \includegraphics[width=0.5\textwidth]{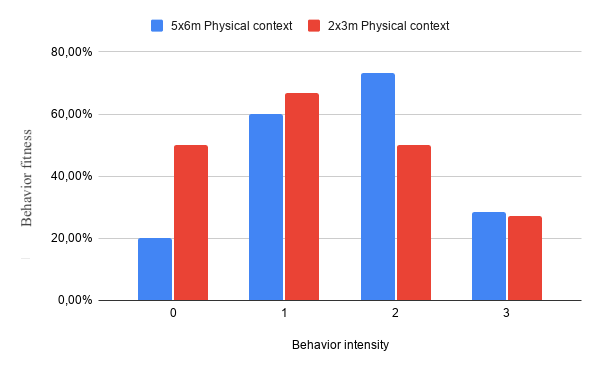} \caption{The calculated fitness of each behavior intensity based on positive ratings for each behavior in the two physical contexts. The blue series depicts the largest (6x5m) physical context, while the red series depicts the smaller (2x3m) physical context. The x-axis shows the different behavior intensity levels while the y-axis shows the calculated fitness of each behavior. }
\label{percentage}
\end{figure}

Figure \ref{percentage} shows the results of the second training
experiment. The graph illustrates the robot's calculated behavior
fitness for each physical context based on the positive feedback
gathered from users in the human-robot interactions for each behavior in
the two training contexts. There is a significant (p\textless{}.05)
difference in the distribution of votes for the two physical contexts
when the robot asked the users to choose the most fitting of the two
displayed behaviors. The found most fitting behavior for the largest
(6x5m) context was the behavior with intensity level 2 (with positive
feedback in 73\% of times tested), while the lowest-rated behavior from
the same context was the behavior with intensity level 0 which was only
found most fitting in 20\% of the times it was tested. For the smaller
physical context (2x3m) the behavior found to be the best fitting was
the behavior with intensity 1 with 66.7\% positive votes, while the
least fitting behavior was the behavior with intensity level 3 27.3\% of
the positive votes.

\begin{figure}[h] \centering \includegraphics[width=0.5\textwidth]{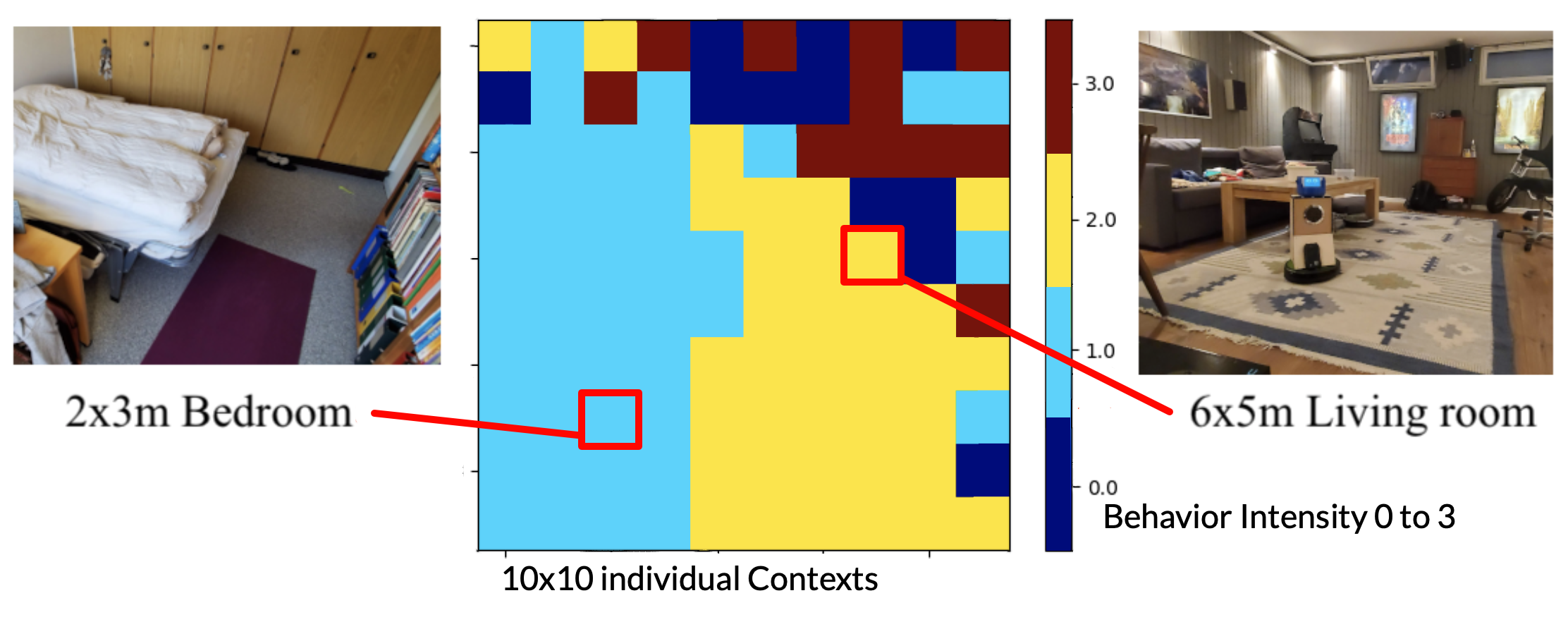} \caption{The context behavior representation. The 10x10 map represents the 100 individual contexts, while each color representing the highest prioritized behavior intensity of each context. The light blue color represents intensity level 1 while yellow color is intensity level 2. The topological positions are similar to those in Figure \ref{context_map}. Note that this map shows the behavior associated with the different contexts, not a map of the physical location.}
\label{behaviors}
\end{figure}

Following both training phases, the robot had fully updated its behavior
prioritization in accordance with each interaction result. The
visualization of top behaviors for each context in the 10x10
context-representation can be seen in Figure \ref{behaviors}. The
visualization depicts the two main physical context regions and the
voted most fitting behavior intensity in each of them with an intensity
level of 2 (yellow) for the 6x5m physical context and an intensity level
of 1 (light blue) for the 2x3m physical context. The maximum intensity
level of 3 is found mainly in the upper right region of the context
representation which matches the largest found physical context
measurements in Figure \ref{context_map}. Comparing the physical context
visualization and the context behavior representation reveals that the
adjusted behavior intensity levels intuitively match the physical
aspects of each context but also reveals that the measurements performed
before each interaction often place the robot between two well-defined
context regions in the representation.

\subsection{Validating the trained framework}

The validation experiment included 90 participants. Out of those
participants, 58 preferred the behavior identified by the robot as being
the optimal behavior for the physical context while 32 participants
preferred a different randomly selected behavior instead. A binominal
test highlight that it is a significant result with the one-tailed
probability of exactly, or greater than, k=58 out of n=90 is p =
.004204. There was a slight tendency that the average Likert ratings of
the framework-selected behavior were higher than the average ratings of
the other three behaviors, but the experiment lacked sufficient
participants to provide a significant result to prove that beyond doubt.

Summing up the results there were two main findings. The first finding
is that the framework managed to find a sound prioritization for
behavior intensity by adjusting the behavior prioritization for each
context in accordance with the physical attributes of each context. The
second finding is that the trained framework, when put in a never before
encountered physical context, managed to identify the best possible
behavior for the physical circumstances verified to a significant extent
by human observers.

\section{Applying the findings}

This paper aimed to construct a robot that utilized simple physical
context information as a guide to drive its behavior selection. There is
a definite and clear relationship between the physical properties of the
room and the human-preferred physical properties of the robot's
behaviors. The results indicate that as the physical space increases
around the robot, and the participants who interact with it, so does the
amount of preferred intensity in the robot's behaviors. The robot
managed to create regions in its context representation for each
physical context and update the same regions with behavior that matched
the increased physical dimensions. However, Affecta V3 did not map the
context consistently to the same node in the context representation.
This stems from the uncertainty of using a single attribute to recognize
the current context. Adding further parameters to identify the physical
aspects of the contexts would make it more precise. However, there is a
strength and a point in using a single attribute, as it is easy to apply
to most robots and it provides contextual information even from simples
sensors.

The robot prioritized between discrete predefined behaviors using
physical context attributes which admittedly is an extreme
simplification of complex psychological processes. As it is making
assumptions on a sparse set of information it will often be wrong when
determining contexts and best behaviors. However, the same can be said
for most humans. We don't always find the completely correct behavior
for a given situation, we adjust our immediate behaviors and negotiate
affective status many times through each interaction
\autocite{doi:10.1177/1557234X11410385}. More complex and richer sensors
could provide better distinctions between contexts but may also
introduce more sensor noise. Some attributes aid behavior selection
better than others. E.g the detected color of various rooms would make
it easy to recognize a known context, but that attribute says nothing
about the optimal behaviors. The kind of behaviors that the robot
optimizes for determines the choice of attributes, meaning that it is a
tradeoff between high-resolution context distinction and more
generalizable behaviors.

The last two behaviors with intensity 0 and 3 were mostly
down-prioritized. Outside of them not generally fitting the context,
this could also be due to the nature of the interaction. The
participants interacted with the robot from a distance between 1-2
meters and a large physical behavior (intensity 3) might seem
intimidating coming from a robot, and a lack of movement (intensity 0)
might make the robot seem inanimate or uninteresting. However, some
people preferred these behaviors and it indicates that personal
preference also plays a part in each interaction. The individual mood
and personality of each test participant may also influence how each
behavior is received.

\section{Conclusion}

This paper investigated the abilities of a behavior-prioritization
framework that adapts behaviors to the physical contexts. The framework
enabled a robot to autonomously prioritize between predefined behaviors
in each encountered context. The framework was trained in two different
physical contexts through 72 interactions with 6 human participants. The
ability to distinguish contexts was manually verified as working
successfully. The result showed that the robot created distinguishable
regions in the visualized context-representation for each visited
physical context. The framework also managed to autonomously prioritize
behaviors in each identified context. The resulting list of behaviors
matching each physical context was learned through interactions with the
users, and the calculated fitness of each behavior and the visualized
map of behaviors showed that the corresponding behaviors had been
correctly prioritized by the robot. The trained framework was verified
in an experiment that showed the framework-chosen behavior was the best
possible match for a new context as chosen by human observers to a
significant extent. The Affecta-context framework highlight the
potential in context-based behavior selection and demonstrate the
possibility of basing context-based behavior selection on the
information retrieved from ad-hoc measurable attributes during an
interaction. Future breeds of social robots should adapt to contextual
information and this paper arguest that while a richer context
distinction may be preferred for context-guided robot behavior
selection, even simple attributes can provide the foundation for context
informed behavior selection.

\balance
\bibliography{main}
\bibliographystyle{IEEEtran}

\end{document}